\documentclass[a4paper,twoside]{article}

\usepackage{epsfig}
\usepackage{subcaption}
\usepackage{calc}
\usepackage{amssymb}
\usepackage{amstext}
\usepackage{amsmath}
\usepackage{amsthm}
\usepackage{multicol}
\usepackage{pslatex}
\usepackage{apalike}
\usepackage{hyperref}
\usepackage{SCITEPRESS}     % Please add other packages that you may need BEFORE the SCITEPRESS.sty package.

\begin{document}

\title{Altering Facial Expression Based on Textual Emotion}

\author{\authorname{Mohammad Imrul Jubair\sup{1},
Md. Masud Rana\sup{1},
Md. Amir Hamza, 
Mohsena Ashraf,\\
Fahim Ahsan Khan,
Ahnaf Tahseen Prince
}
\affiliation{Department of Computer Science and Engineering, Ahsanullah University of Science and Technology, Bangladesh}
% \affiliation{\sup{2}Department of Computing, Main University, MySecondTown, MyCountry}
\email{\{jubair.cse, mohsena\_ria.cse\}@aust.edu, \{masud.cseian, hamza.cseian, fahimahsankhan, ahnaftahseen\}@gmail.com}
}

\keywords{Facial Expression, Image to Image Translation, Emotion Detection}

\abstract{
% Digital pictures of facial emotions are one of the most powerful subjects. While extracting emotions from photos is a well-established job in the field of computer vision, doing the opposite—-synthesizing face expressions in photographs—-is relatively new. This procedure of
% % producing pictures with alternative facial expressions or 
% % changing an existing expression in an image necessitates the use of the Generative Adversarial Network (GAN).
% The purpose of this paper is to demonstrate how to modify a person's facial expression in a picture using GAN, where the input image with an existing emotion (i.e. joyful) is transformed to a new expression (i.e. disgust).
Faces and their expressions are one of the potent subjects for digital images. Detecting emotions from images is an ancient task in the field of computer vision; however, performing its reverse---synthesizing facial expressions from images---is quite new. Such operations of regenerating images with different facial expressions, or altering an existing expression in an image require the Generative Adversarial Network (GAN). In this paper, we aim to change the facial expression in an image using GAN, where the input image with an initial expression (i.e., happy) is altered to a different expression (i.e., disgusted) for the same person. 
We used StarGAN techniques on a modified version of the MUG dataset to accomplish this objective.
% We expanded our work by redesigning facial expressions in an image based on the mood conveyed in a specific text.
Moreover, we extended our work further by remodeling facial expressions in an image indicated by the emotion from a given text.
As a result, we applied a Long Short-Term Memory (LSTM) method to extract emotion from the text and forwarded it to our expression-altering module. As a demonstration of our working pipeline, we also create an application prototype of a blog that regenerates the profile picture with different expressions based on the user's textual emotion.}

\maketitle

\section{\uppercase{Introduction}}
\label{sec:introduction}
\footnotetext[1]{These authors contributed equally to this work.}
As a result of the widespread usage of social media and blogs, people have been accustomed to expressing their feelings and thoughts digitally, whether by text, voice, or image. When it comes to these emotions, the facial expression plays an important role in our everyday modes of communication and connection, particularly when it comes to pictures, videos, online conferences, etc. While speaking of facial expressions, we commonly refer to happiness, sadness, anger, disgust, etc, which are very natural for humans~\cite{psyg1}\cite{FRANK20015230}\cite{fpsyg}.

Photos and their expressions have an undoubtedly strong impact in the field of computer vision since researches have been going on for years to extract expressions. The recent progress of machine learning has brought enough accuracy in detecting and recognizing facial expressions. It is not simple to tackle the opposite difficulty of this task---imposing varied emotions on a pre-existing face in a photograph without the help of another human being—--and this subject is still mostly studied. The recent breakthrough of the Generative Adversarial Network~\cite{goodfellow2014generative} has influenced the researchers to work with image-to-image translations and to develop different stunning tools, for example converting a horse into a zebra~\cite{CycleGAN2017}. This type of application of GAN typically converts the image from a source domain $X$ to a target domain $Y$ by learning from an adequate amount of image data.
% In light of the fact that generative models~\cite{generative_model} 
Hence it also make it possible to create new images by learning from a large dataset, they are extremely appealing tools for creating images with a variety of expressions. When applied to an input image, it may be used to transform the facial expression into the intended expression; for instance, from a joyful face to an angry face. A dataset of faces of diverse people with a range of expressions is required for this type of modification.

\subsection{Contribution}
We found this research domain of regenerating different facial expressions very exciting and, in this paper, we experimented on different GAN-based methods on a variety of datasets as an attempt to determine the fittest combination. To make our research even more interesting, we broadened the scope of our domain to include an application that went beyond just modifying the facial expression; we concentrated on textual emotion transmission to images. In our method, there are two phases: first, we extract emotion from a text, and then we transmit the outcome as an input to the facial expression generating module, which then modifies the emotion in a person's photograph.

In real-world situations---such as blogs and social media---our method can be utilized. As an example, consider a blog where the facial expression of a user's photo may be modified based on the content of his or her most recent post. For instances, in the case of a sorrowful statement like ``\texttt{I'm not feeling well today}'', the expression on his already-existing profile photo---which had a happy face---would instantly change to one of sadness. Our approach requires two inputs: one is a photograph of the person's face and the other is the text of the person's post. The image is then sent through an artificial neural network module to be translated into yet another image of the same person but with a different expression based on the emotion collected from a text. In spite of this, our suggested text-to-image emotion transmission pipeline may be implemented into any instant messaging program, where it will identify emotions from the discussion and create expressions on thumbnail images of the people being spoken with. For example, if there is some talking that may contain something unpleasant or linked to harassment, one's photo may be changed out of disgust or rage.\\

Contributions of this paper are summarized below. 
\begin{itemize}
    \item We provide brief explanations of the image datasets of face expression for facial emotion creation, which can serve as a useful reference for future studies. We experimented with various GAN models on picture datasets in order to create faces with the required emotion, and in this paper, we highlight the outcomes for further investigation. Furthermore, we make required adjustments to the datasets in order to get better results. We used Long Short-Term Memory (LSTM) model \cite{hochreiter1997long} to train for the task of extracting emotion from text. The retrieved emotions are fed into our facial expression creation algorithm, which then generates facial expressions.
    \item We provide the findings of our suggested pipeline as well as a prototype application to illustrate our point of view. We developed a blog where the user can upload a post and our model first detects the emotion of the post and then apply the emotion over his/her face, and generates the expression corresponding to the emotion.
\end{itemize}

\textbf{Paper Organization. }
The following describes the structure of this paper. Section~\ref{sec:related} discusses related studies on face expression creation as well as relevant datasets in more detail. In Section~\ref{sec:approach}, we describe our suggested pipeline and methodology for our image emotion transfer from text, and in Section~\ref{sec:exp}, we describe the outcomes of our experiment. Section~\ref{sec:con} concludes our discussion by outlining the limits of our research as well as possible future directions.

\section{RELATED WORKS}
\label{sec:related}
Various relevant studies on GAN for facial expression genetation are discussed in this section. We also explore several picture datasets of face expressions, which are subsequently followed by a number of text-based datasets for the purpose of emotion recognition.

\subsection{\textbf{Facial Expression Generation}}
Generative Adversarial Networks (GANs)~\cite{goodfellow2014generative} were used in our research to generate pictures with a variety of face emotions. GAN is an adversarial method that is comprised of two neural network models: the generator and the discriminator.
The generator model attempts to learn the data distribution, while the discriminator model attempts to differentiate between samples taken from the generator and samples taken from the original data distribution. During the training process, these two models are trained in parallel, with the generator learning to create more and more realistic examples while the discriminator learns to become more and more accurate at differentiating produced data from actual data. As a continuous game, both networks strive to make the produced samples seem as indistinguishable from actual data as possible. The loss function of GAN can be represented by the following equation where the generator tries to minimize it and the discriminator tries to maximize it. The loss function of a GAN model is shown in Eq.~\ref{eq:ganloss}.
\begin{equation}
\label{eq:ganloss}
    L (G, D) = E_{x}[\log (D(x))]+E_{y}[\log (1-D(G(y)))]
\end{equation}
Here, $x$ is the real data sample where $E_{x}$ is the expected value over all $x$. $D(x)$ is the discriminator's probability estimation of $x$ being real, and $D(G(y)$ is the discriminator's probability estimation that a fake instance $y$ is real.

\subsubsection{Study on Different Methods}
In recent years, a large number of variants of the Generative Adversarial Network have been developed. Various kinds of GANs have been employed to produce face pictures with specific expressions by a variety of researchers, including several types of Convolutional Neural Networks~\cite{CNNpaper}. We've compiled a list of some of the more noteworthy ones.

\begin{itemize}
    \item \textit{ExprGAN~\cite{exprgan}:} The Expression Generative Adversarial Network (ExprGAN) is a neural network that generates random and low-resolution pictures of faces using an input face image and a labeled expression. With the face picture and their emotion label, it trains the encoder to produce fresh photos of the same person's face with a different expression. With the assistance of this model, the strength of the produced facial expressions may be adjusted from high to low.
    
    \item \textit{StarGAN \cite{stargan}:} An image-to-image translation method known as Star Generative Adversarial Network (StarGAN) produces fixed input facial expressions for various domains based on a single input face. It is capable of learning mappings across domains using just a single generator and a discriminator, which makes it very efficient. The majority of the work on this model was done using CycleGAN, which is a tool for transferring pictures from one domain to another. As previously stated, StarGAN is comprised of two convolutional layers, with the generator using instance normalization and the discriminator employing no normalization. In addition to the PatchGAN~\cite{isola2016imagetoimage} discriminator network, which determines whether local image patches are genuine or false, the StarGAN discriminator network is based on the Discriminator network of StarGAN. However, it will not be able to create a face emotion that is not already included in the training set.
    
    \item \textit{G2GAN~\cite{song2018geometry}:} In the training phase of the Geometry-Guided Generative Adversarial Network (G2GAN), a pair of GANs is used to execute two opposing tasks, which are performed by the G2GAN. One method is to eliminate the expressions from face pictures, while another is to create synthesized expressions from facial photographs. In conjunction with one another, these two networks form a mapping cycle between the neutral facial expression and the random facial expressions on the face. The face geometry is used to regulate the synthesis of facial expressions in this method of control. Additionally, it maintains the individuality of the expressions when synthesizing them.
    
    \item \textit{CDAAE~\cite{zhou2017photorealistic}:} The Conditional Difference Adversarial Autoencoder (CDAAE) produces synthetic facial pictures of a previously unknown individual with a desired expression based on the conditional difference between the two images. While learning high-level facial expressions, CDAAE uses a long-range feedforward connection that runs from the encoder layer to the decoder layer, and it only takes into consideration low-level face characteristics while learning high-level facial emotions. Instead of using the same pictures as input and output, the network is trained using pairs of photographs of the same person with different expressions rather than using the same images as input and output. This method maintains the identity of the data and is appropriate for use with even smaller datasets.
    
    \item \textit{A Text-Based Chat System Embodied with an Expressive Agent~\cite{Alam2017}:} Here the author proposes a framework for a text-based chat system with a life-like virtual agent that seeks to facilitate natural user interaction. They created an agent that can generate nonverbal communications like facial expressions and movements by studying users' text messages. This agent can generate facial expressions for six fundamental emotions: happy, sad, fear, furious, surprised, and disgust, plus two more: irony and determination. To depict expressiveness, the authors used the software programs---\textit{MakeHuman} and \textit{Blender}---to build two 3D human characters, a male and a female and to create realistic face expressions for these agents. Instead than modifying the user's picture, the writers utilized an animated figure to convey emotions.
\end{itemize}

There are, however, different kinds of GAN models that may be used for the creation of face expressions. In the papers \cite{deng2019cgan} and \cite{liu2021self}, the authors used conditional GAN (cGAN) for the generation of $7$ expressions (anger, disgust, fear, happy, sad, surprise, and neutral) and $6$ expressions (anger, disgust, fear, surprise, sadness, and happiness), respectively. \cite{chen2018double} used Double Encoder Conditional GAN (DECGAN) to generate seven different expressions in a single run. A further development is the Geometry—Contrastive Adversarial Network (GC-GAN), which was developed by \cite{qiao2018geometry} for the generation of face pictures with target expressions. But none of these methods took into account the possibility of creating face pictures with emotions from text.
As a result, in order to achieve our goal of emotion transmission to image from text, we concentrated on extracting emotions from the text and then creating face expressions by combining those feelings with others.

\subsubsection{Study on Facial Expression Datasets}
% A key issue for any kind of work involving computer vision is the selection of data sets to be used.
Several kinds of datasets were explored for generating facial expressions and generating emotion from text. The datasets that were utilized for the creation of face expressions are listed below.

\begin{itemize}
    \item \textit{CelebFaces Attributes Dataset (CelebA)~\cite{celebA}:} It is a large-scale face attributes dataset including more than $200$K celebrity pictures, each of which has $40$ attribute annotations. It is available for download here. The pictures in this collection depict a wide range of posture variations as well as a cluttered backdrop.
    % Among the many diverse elements are huge amounts of information and detailed annotations.
    There are over $10,000$ identities, over $202$ thousand facial pictures, and five landmark locations with $40$ binary characteristics annotations each image.
    
    \item \textit{Multimedia Understanding Group (MUG) \cite{mug}:} In order to address some of the constraints of previous comparable databases, such as high resolution, consistent lighting, a large number of subjects and several takes per subject, the MUG database was developed. It is made up of picture sequences of $86$ people expressing themselves via facial expressions.
    % The individuals were seated on a chair in front of a camera for the photograph.
    % The camera used to collect the pictures for this dataset was capable of capturing images at a rate of $19$ frames per second; 
    Each image was recorded in the jpg format with a resolution of $896\times896$ pixels. There were $35$ women and $51$ men who took part in the database creation.
    % all of them are of Caucasian descent and vary in age from $20$ to $35$ years.
    The participants were divided into two groups: women and men.
    With $1462$ sequences accessible, each including more than $1$ thousand pictures and seven different face expressions, the possibilities are endless.
    The reactions range from surprise to delight to fear to rage to neutrality to sorrow to contempt.
    
    \item \textit{Facial Expression Research Group Database (FERG) \cite{ferg-paper-bib}:} It is mostly a 2D animation dataset that contains pictures of six stylised characters' facial expressions, most of which are animated ($3$ males and $3$ females). Annotated facial expressions are used to create a database of stylised characters in the game. It includes approximately $55$K annotated face pictures of six styled characters, which are organized into categories. The characters were created using the \textit{MAYA} program and have six distinct face emotions: furious, disgusted, fear, delight, surprise, neutral, and sad.
    % The characters were created using the \textit{MAYA} software and have six different facial expressions.
    
    \item \textit{Oulu-CASIA NIR-VIS Database (Oulu CASIA) \cite{oulu}:} Approximately $74\%$ of the participants in the Oulu-CASIA database are male, with a total of $80$ subjects ranging in age from $23$ to $58$ years old in the database. Fifty topics are Finnish, and thirty subjects are Chinese, according to the course description. Images are captured at a rate of $25$ frames per second and at a resolution of $320$ by $240$ pixels by using imaging gear.
    % A total of three distinct lighting settings were used to capture all of the emotions in the pictures in this dataset: normal, weak, and completely dark.
    % A total of three parts are included in this dataset: the original films, JPEG pictures of the original movies, and JPET images of cropped faces.
    In the database, there are six different face expressions to choose from: surprise, happiness, sorrow, rage, fear and contempt.
    
    \item \textit{AffectNet Database~\cite{affectnet}:} Developed via the collection and annotation of face pictures, AffectNet is a library of naturalistic facial emotions. There are almost one million face pictures in this collection, which was compiled from the Internet by searching three major search engines with $1,250$ emotion-related keywords in six different languages. Manual labeling was performed on about half of the recovered pictures ($440$K), which included seven different facial expressions and their respective intensities in terms of valence, arousal, and agitation. It includes seven different face expressions: anger, contempt, fear, joy, neutrality, and sorrow.
\end{itemize}
\begin{figure}[ht]
    \centering
    \includegraphics[width=0.85\linewidth]{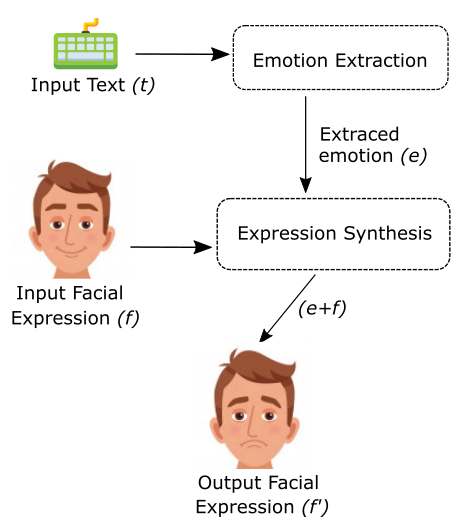}
    \caption{The proposed pipeline of our approach.}
    \label{fig:method}
\end{figure}
\begin{table}[h]
\centering
\begin{tabular}{cc}
\hline
\textbf{Expressions} & \textbf{\# of entries} \\ \hline \hline
happy                    & $1092$                   \\ \hline
sadness                  & $1082$                   \\ \hline
anger                    & $1079$                   \\ \hline
fear                     & $1076$                   \\ \hline
shame                    & $1071$                   \\ \hline
disgust                  & $1066$                   \\ \hline
surprise                 & $1050$                   \\ \hline
\textbf{Total}           & \textbf{$7516$}          \\ \hline
\end{tabular}
\caption{A statistical overview of EmoBank dataset \cite{emobank}.}
\label{tab:my-table}
\end{table}
\subsection{\textbf{Dataset for Emotion Detection from Text}}
In order to extract the emotion from the text, we utilized the EmoBank Dataset~\cite{emobank}. A text corpus of emotion is available, including texts gathered from different social media platforms and the internet as a whole. Emotion categories are manually assigned to each text corpus in a process called manual labeling. Seven different kinds of emotions are represented by a total of 7,516 items, including joy, sorrow, rage, fear, humiliation, disgust, and feeling guilty.
Table~\ref{tab:my-table} shows the total number of entries for each emotion in this dataset.

We gathered and analyzed a variety of datasets and used a variety of techniques to these datasets while maintaining the same experimental setup in order to determine the most appropriate combination. The (LSTM+EmoBank) combination was chosen in this study to identify emotional content in text since it fulfills our goal in a simple way. As a side note, we discovered that StarGAN on a tweaked version of the MUG dataset performs the best for us. After describing our pipeline, which is based on the previously stated blended approach, we will go through the experimental setups and comparisons in the next part.
 \begin{figure}[ht]
\centering
\begin{subfigure}[b]{.2\textwidth}
  \centering
  \includegraphics[width=\textwidth]{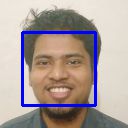}
  \caption{}
  \label{fig:haarbbox}
\end{subfigure}
% \hfill
\begin{subfigure}[b]{.12\textwidth}
  \centering
  \includegraphics[width=\textwidth]{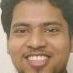}
  \caption{}
  \label{fig:facecrop}
\end{subfigure}
\caption{Results of applying Haar-cascade. \textit{(a)} main input image, and \textit{(b)} shows final cropped and resized image.}
\label{fig:haarcascade}
\end{figure}
 \begin{figure}[ht]
\centering
\begin{subfigure}[b]{.3\textwidth}
  \centering
  \includegraphics[width=\textwidth]{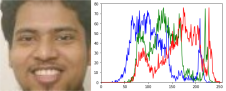}
  \caption{} \vspace{1mm}
  \label{fig:histin}
\end{subfigure}
% \hfill
\begin{subfigure}[b]{.3\textwidth}
  \centering
  \includegraphics[width=\textwidth]{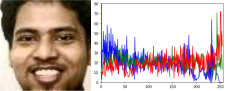}
  \caption{}
  \label{fig:histout}
\end{subfigure}
\caption{Results of applying histogram equalization. \textit{(a)} input image of the face \textit{(left)} with its histogram \textit{(right)}, and \textit{(b)} output image of the face \textit{(left)} with its histogram \textit{(right)} after the equalization.}
\label{fig:hist}
\end{figure}
\section{PROPOSED PIPELINE}
\label{sec:approach}
The goal of our work is to change the facial expression of a photograph depending on the emotion derived from a particular text. The pipeline for our system is shown in Fig.~\ref{fig:method}. There are two stages to the pipeline's operation. In the beginning, it accepts the text $t$ and the first face picture that is entered.
To identify the emotion $e$, the text is delivered to the emotion extraction module, and the picture of the facial expression $f$ is provided to the expression synthesis module. An image of the person's face $f'$ is produced using an expression based on the data in the synthesis phase $(f+e)$ during the synthesis phase. Following that, we will go through each of these stages in more detail.

\subsection{\textbf{Emotion Extraction from Text} (LSTM+EmoBank)}
The EmoBank dataset~\cite{emobank} was used in conjunction with Long Short Term Memory to aid in the emotion identification process~\cite{rnn_lstm}. In order to improve adaption, we do preprocessing on the dataset, which includes case conversion, removal of white space and punctuation marks, spell correction, and handling of numerical symbols and the unknown term, among other things. Afterwards, we embed the text using the GloVe~\cite{glove_pennington} representation method and train our model to recognize emotions in the text provided by the user.

\subsection{\textbf{Facial Expression Synthesis} (StarGAN + 
\textit{tunedMUG})}
We use the StarGAN\cite{stargan} to change the facial expression of a person depending on the emotion expressed in a text message. In order to test the technique, we used a modified version of the MUG dataset~\cite{mug}, which we refer to as the \textit{tunedMUG} dataset. The actions that were taken in order to acquire this version are detailed below.

\subsubsection{Face Extraction} Generalization is required in order to run a model across any kind of data. Face expressions need data from a variety of backgrounds, individuals, and situations to be accurate. In order to construct a generalized model, we first applied the Haar-cascade~\cite{harcascade_inproceedings} to the MUG dataset, which allowed us to concentrate on the faces as much as possible throughout the training process. Thousands of positive pictures (such as photos of faces) and thousands of negative images are used to train the Haar-cascade (images without faces). It provided us the location of the faces and we stored this face as an image of $128\times 128\times 3$ size. Fig.~\ref{fig:haarcascade} shows the output of applying the Haar-cascade.

\subsubsection{Histogram Equalization}
The Histogram Equalization method~\cite{CHENG2004158} is used to ensure that all of the picture data has the same distribution. The samples in the dataset are given a generic attribute as a result of this technique. The outcome of the procedure is shown in Fig.~\ref{fig:hist}.
\begin{figure}[h]
    \centering
    \includegraphics[width=\linewidth]{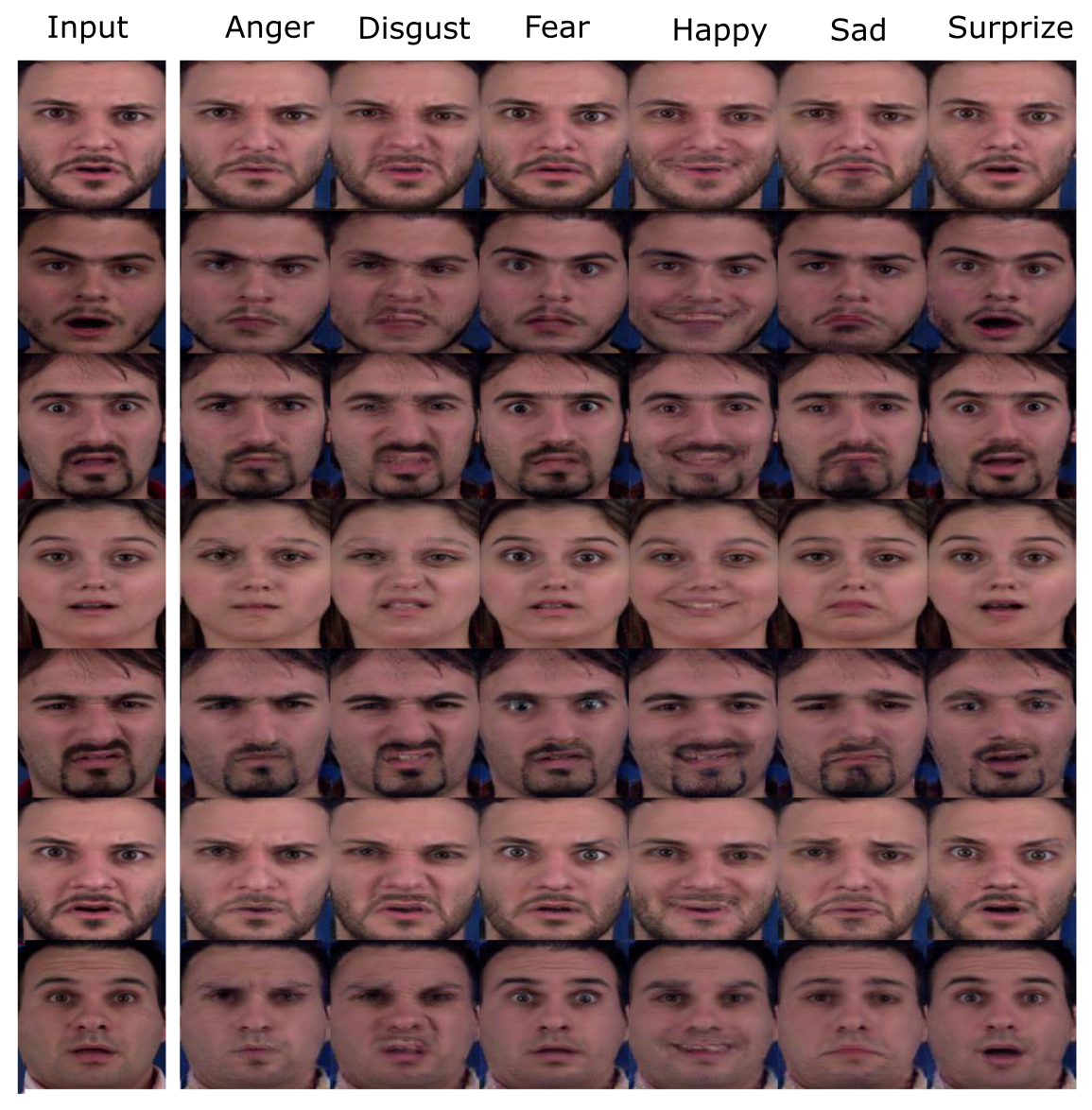}
    \caption{Results of facial expression synthesis for test images. Here the, images in the left column are the input images, and the other columns represent different expressions.}
    \label{fig:results2}
\end{figure}
\begin{figure}[htb]
    \centering
    \includegraphics[width=\linewidth]{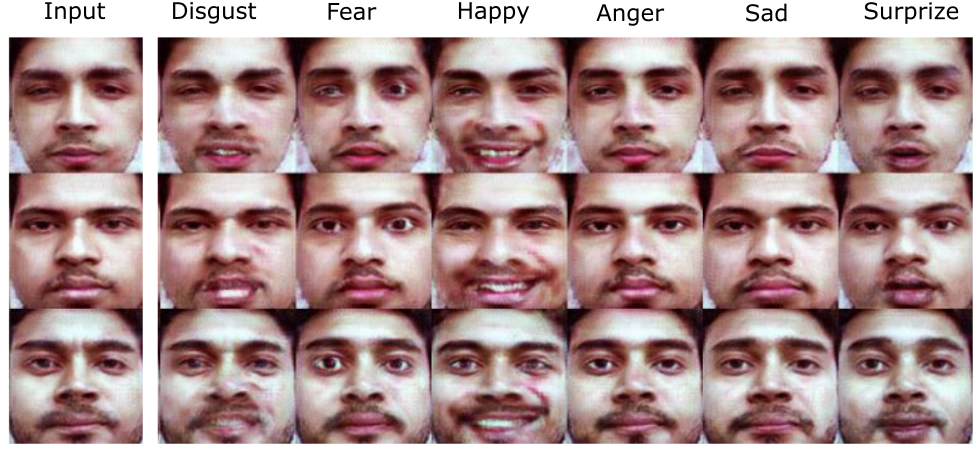}
    \caption{Additional results of facial expression synthesis of our pipeline for test images. Here the faces are of different sources than the original MUG dataset.}
    \label{fig:results}
\end{figure}

\section{EXPERIMENTS}
\label{sec:exp}
As part of our pipeline development, we use the (StarGAN + \textit{tunedMUG}) model and the (LSTM + EmoBanK) model.
% where the first model identifies textual emotion and passes this information along to the latter model.
In this part, we describe our findings in suitable depth and with relevant comparisons.

\subsection{\textbf{Experimental Setup}}
With $11.17$GB GPU support, we were able to put the StarGAN model into action on a \texttt{Google Colab}. It took more than $48$ hours to complete our whole training process from start to finish. We utilized pictures from our \textit{tunedMUG} dataset worth about $20$k for training purposes. We utilized the Adam optimizer~\cite{kingma2017adam} with decay rates of $\beta_1 = 0.5$ and $\beta_2 = 0.999$ for the $1^{st}$ and $2^{nd}$ moments of the gradient, respectively, with $\beta_1 = 0.5$ and $\beta_2 = 0.999$ for the $1^{st}$ and $2^{nd}$ moments of the gradient. By flipping data horizontally, we can also add data augmentation into the equation. Furthermore, we choose batch sizes of $16$ and $0.0001$ as the learning rates for the generator and discriminator, respectively, to get the best results.
 \begin{table}[h]
\centering
\begin{tabular}{c c c} 
 \hline
  \textbf{Model+Dataset}
  &
  \textbf{Train Acc.}
  & \textbf{Test Acc.}\\
 \hline \hline
 RNN + EmoBank & 44\% & 33\% \\
 \hline
 LSTM + EmoBank &  71\% & 59\% \\
 \hline
\end{tabular}
\caption{Comparison of accuracy of RNN and LSTM based emotion extraction from text on EmoBank dataset.}
\label{table:emoBank_table}
\end{table}
In addition, we base our LSTM model on the \texttt{Google Colab}. To train our model, we created an embedding matrix using our dataset and used the \texttt{glove.6B.50d}~\cite{lal_2018}, which improved the consistency of our model by identifying similar words in our dataset and included them in our embedding matrix.
This embedding matrix assists the model in dealing with the user's input term that was not in our dataset, and the model identified a comparable word to this unknown word in order to accurately generate the outcome. LSTM was used with $\beta_1=0.9$ and $\beta_2=0.999$ for Adam optimizer and categorical cross-entropy \cite{zhang2018generalized} as the loss function
% (Eq.~\ref{eq:cross})
with a batch size of $32$.
% \begin{equation}
% \label{eq:cross}
%     L_c(\hat{y},y) =  -\frac{1}{N} \sum_{i}^{N}
%      [y_{i}\log\hat{y}_{i} + (1-y_{i})\log(1-\hat{y}_{i})]. 
% \end{equation}
% Here, $N$ is the number of samples, $y$ is the probability distribution of true labels and $\hat{y}$ is the probability distribution of the predicted labels.
\begin{figure}[ht]
    \centering
    \includegraphics[width=\linewidth]{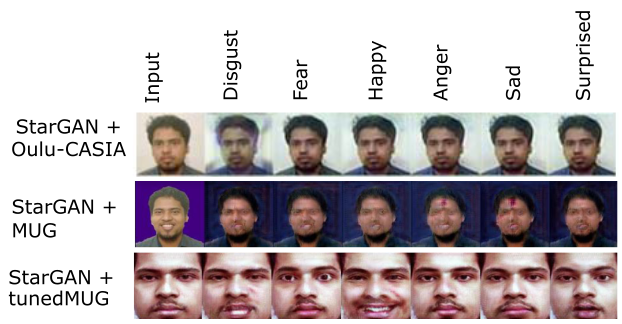}
    \caption{Comparison among Oulu-CASIA \textit{(top)}, MUG \textit{(middle)} and our tunedMUG dataset \textit{(bottom)} for applying StartGAN. Here, the leftmost column holds the input faces.}
    \label{fig:comparison}
\end{figure}

\subsection{\textbf{Experimental Results}}
With the EmoBank datset, we ran tests on it with a Recurrent Neural Network (RNN)~\cite{rnn} and a Long Short-Term Memory (LSTM)~\cite{rnn_lstm}, and discovered that LSTM provided acceptable results in terms of test and train accuracy (see Table~\ref{table:emoBank_table}).

The outputs of the expression synthesis module generated from our \textit{StarGAN+tundedMUG} technique for the testing samples are shown in Fig.~\ref{fig:results}, which includes the results of the tests. We also present the findings for faces other than those from the MUG dataset, which are more diverse (Fig.~\ref{fig:results2}). The findings show that our system is capable of producing acceptable outcomes in terms of expression synthesis, which is encouraging.

We also show in Fig.~\ref{fig:comparison} a qualitative comparison between the Oulu-CASIA, MUG, and our \textit{tunedMUG} datasets for the purpose of using StarGAN. The figure shows that our\textit{ StartGAN+tunedMUG} combination produces much better outcomes than the others.

\subsection{\textbf{Application}} 
\begin{figure}[ht]
    \centering
    \begin{subfigure}[b]{\linewidth}
  \centering
  \includegraphics[width=\linewidth]{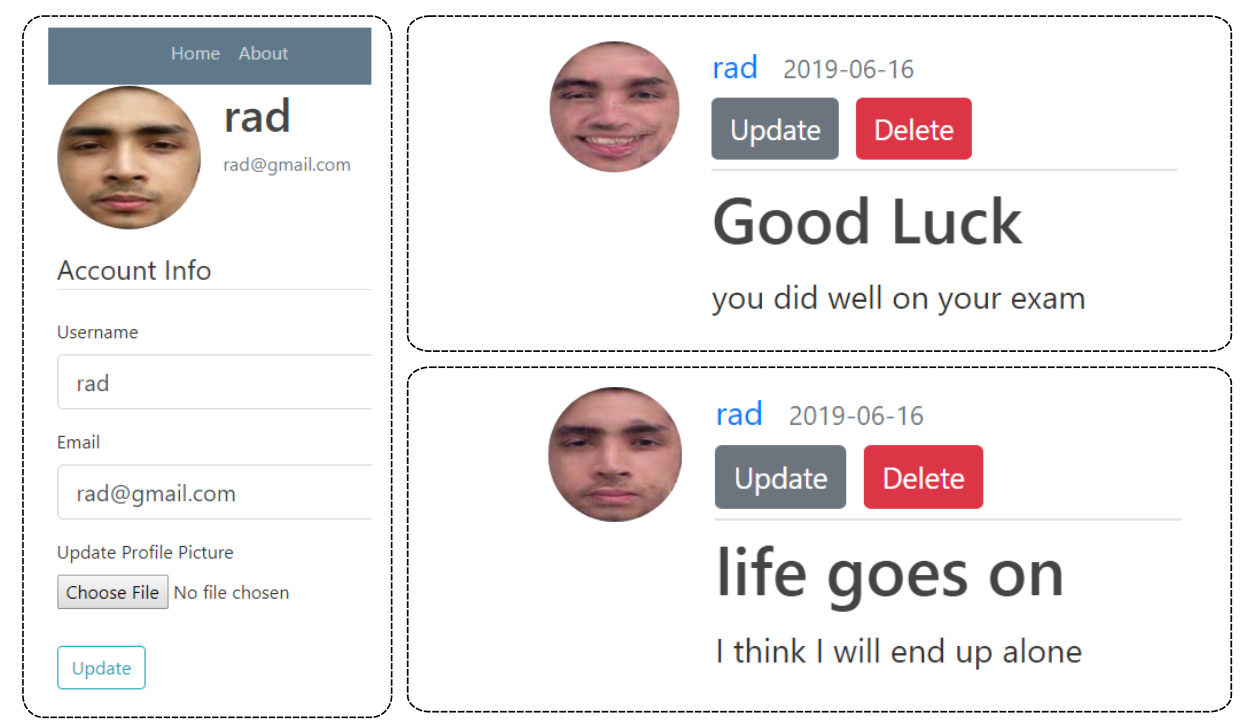}
  \caption{\textit{case - I}}\vspace{1mm}
  \label{fig:app1}
\end{subfigure}
% \hfill
\begin{subfigure}[b]{\linewidth}
  \centering
  \includegraphics[width=\linewidth]{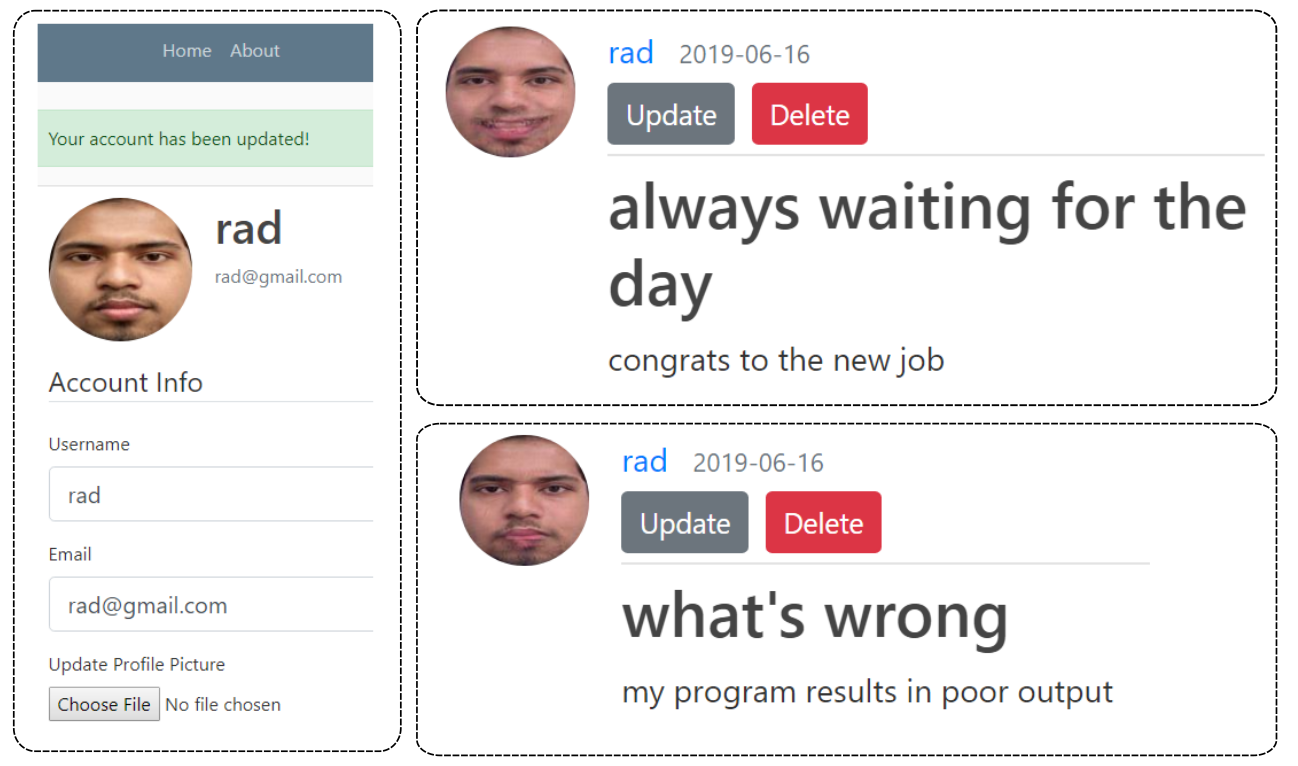}
  \caption{\textit{case - II}} \vspace{1mm}
  \label{fig:app2}
\end{subfigure}
    \caption{Two example cases (a \& b) of using our application. In both cases: \textit{left:} user's profile page with a image of his face. \textit{top--rigth:} user shares a post having a happy emotion and the expression in the profile picture is changed. The similar case occurs in the \textit{bottom--right} but for sadness.}
    \label{fig:whymy}
\end{figure}

We developed a web application in order to demonstrate the overall performance of our image emotion transmission from text method. In this prototype of a social networking site, users can login, add a profile photo, and publish content to the site. Fig.~\ref{fig:whymy} depicts a scenario in which our program is used, in which the user's expression in the image is changed depending on the written post that he has submitted.

\section{\uppercase{Conclusion}}
\label{sec:con}
In this article, we suggested a pipeline for the transmission of emotion from text to picture. Our system receives textual input from the user, extracts emotion from it, and then synthesizes an appropriate facial expression depending on the emotion derived from the text. In order to do this, we divided our system into two phases: one for emotion recognition from text, and another for picture creation using GAN. EmoBank dataset with minimal preparation was utilized for emotion processing, and LSTM was employed to get this result. Based on the MUG dataset, we developed a custom expression synthesis module that can be used in any environment. On this adjusted MUG dataset, we used the StarGAN technique to change the facial expression of the participants. In order to show our working pipeline, we have also created an application that reproduces the profile image with different expressions depending on the mood of the user's post in order to demonstrate our functioning process.\\

There are many possibilities for future endeavors in our system. Currently, our system is focused on the facial area, but we have plans to expand its capabilities to include pictures of the whole body in a variety of positions. In order to achieve better fusion, we want to do ablation studies on more datasets and GAN techniques in the future. We also plan to run user experiment evaluation of our system~\cite{uxmethod}. 

\bibliographystyle{apalike}
{\small
\bibliography{example}}

\end{document}